%% file: final.tex
\documentclass[10pt,twocolumn,letterpaper]{article}

\usepackage{cvpr}
\usepackage{times}
\usepackage{epsfig}
\usepackage{graphicx}
\usepackage{amsmath}
\usepackage{amssymb}
\usepackage{algorithm}
\usepackage{algorithmic}
\usepackage{multirow}
\usepackage{booktabs}
\usepackage[pagebackref=true,breaklinks=true,letterpaper=true,colorlinks,bookmarks=false]{hyperref}

\cvprfinalcopy %

\usepackage[margin=0pt,font=footnotesize,labelfont=bf,labelsep=endash,tableposition=top]{caption}

\def\smallskip{\vspace{0pt}}

\def\cvprPaperID{1290} %

\def\sp{{ ~ }}

\def\subsubsection{\textbf}

\ifcvprfinal\fi
\begin{document}
	\title{Knowledge Adaptation for Efficient Semantic Segmentation\thanks{Accepted to
	Proc.\ IEEE/CVF Conf.\ Computer Vision and Pattern Recognition, USA, 2019.}
	}

	\author{
		Tong He$^1 $         \sp
		Chunhua Shen$^1 $\thanks{Corresponding author: $\tt chunhua.shen@adelaide.edu.au $.}    \sp
		Zhi Tian$^1 $        \sp
		Dong Gong$^1 $       \sp
		Changming Sun$^2 $   \sp
		Youliang Yan$^3 $\\
		$^1 $The University of Adelaide \sp $^2 $Data61, CSIRO \sp $^3 $Noah's Ark Lab, Huawei Technologies
	}

	\maketitle
	\thispagestyle{empty}

	\input{abstract}

	\tableofcontents\clearpage

\input{introduction}

\input{relatedwork}

	\input{methods}

	\input{experiment}

\input{conclusion}
	\input{supplymaterial}

	{\bf Acknowledgements}
	The authors would like to thank Huawei Technologies
	for the donation  of GPU cloud computing resources.

	{\small
		\bibliographystyle{ieee}
		\bibliography{mybib}
	}

\end{document}

%% file: abstract.tex
\begin{abstract}

Both accuracy and efficiency are of significant importance to the task of semantic segmentation. Existing deep FCNs suffer from heavy computations due to a series of high-resolution feature maps for preserving the detailed knowledge in dense estimation. Although reducing the feature map resolution (i.e., applying a large overall stride) via subsampling operations (e.g., pooling and convolution striding) can instantly increase the efficiency, it dramatically decreases the estimation accuracy. To tackle this dilemma, we propose a knowledge distillation method tailored for semantic segmentation to improve the performance of the compact FCNs with large overall stride.
To handle the inconsistency between the features of the student and teacher network, we optimize the feature similarity in a transferred latent domain formulated by utilizing a pre-trained autoencoder.
Moreover, an affinity distillation module is proposed to capture the long-range dependency by calculating the non-local interactions across the whole image.
To validate the effectiveness of our proposed method, extensive experiments have been conducted on three popular benchmarks: Pascal VOC, Cityscapes and Pascal Context. Built
upon a highly competitive baseline, our proposed method can improve the performance of a student network by 2.5\% (mIOU boosts from 70.2 to 72.7 on the cityscapes test set) and can train a better compact model with only 8\% float operations (FLOPS) of a model that achieves comparable performances.

\end{abstract}

%% file: introduction.tex
\section{Introduction}

\begin{figure}
	\centering 
	\includegraphics[width=8cm]{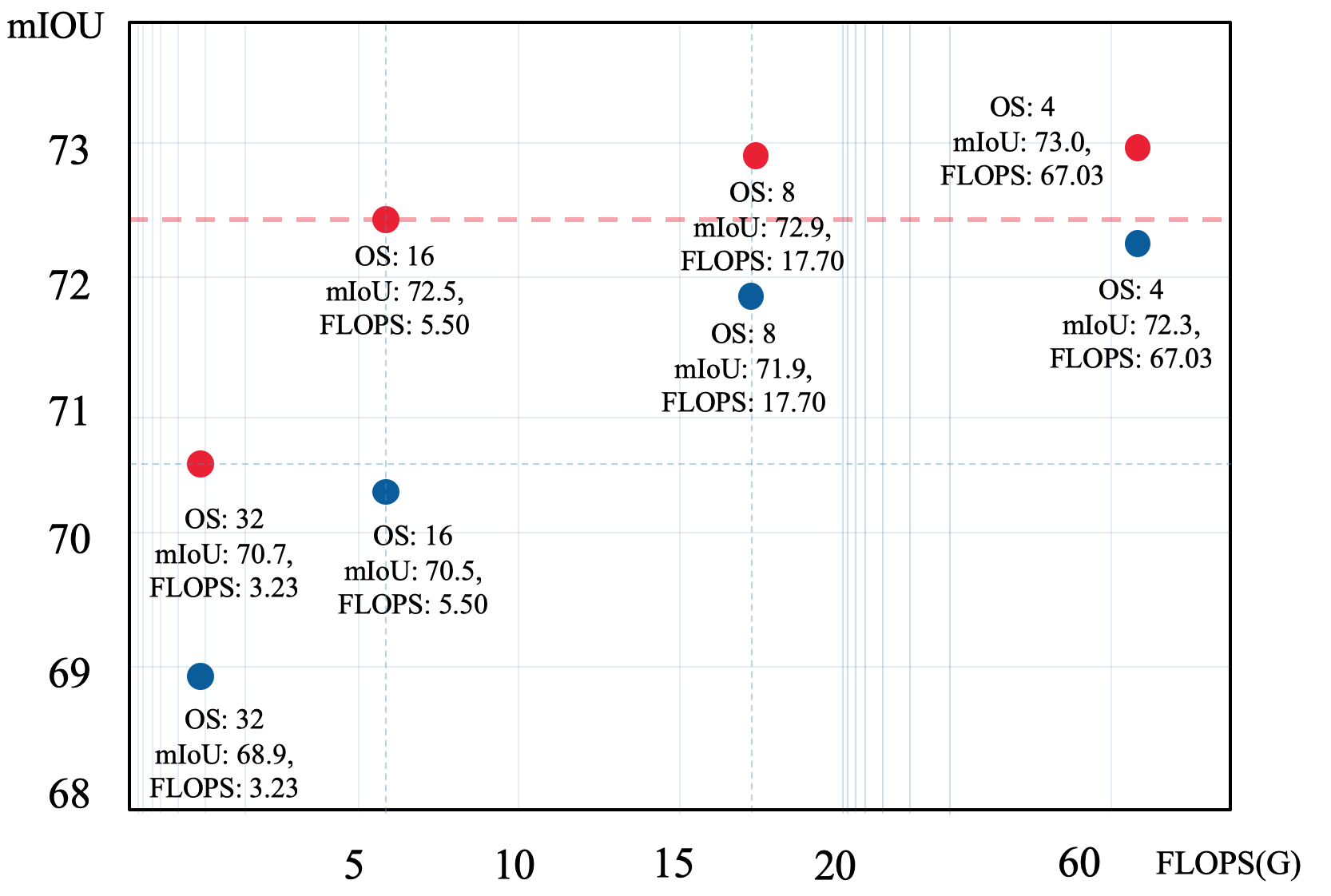} 
	
	\caption{The relation between FLOPS and performance. Blue dots are the performance of student model (MobilNetV2), while red dots are the performance of student model with our proposed knowledge distillation method. The performance is trained on the PASCAL VOC \textit{trainaug} set and tested on the \textit{val} set. \textbf{OS} means output stride. With the help of our proposed method, the student model with low resolution (16s) of the feature maps outperforms the model with large feature maps (4s) by using only 8\% FLOPS.}
	\label{fig:flops}
\end{figure}

Semantic segmentation is a crucial and challenging task for image understanding \cite{Chen2017deeplab, Chen2018deeplabv3plus, Chen2017deeplabv3, Zhao2017psp, Lin2017refinenet, Zhao2017psp, Liu2015semantic, Long2015fully}. It aims to predict a dense labeling map for the input image, which assigns each pixel a unique category label. Semantic segmentation has shown great potential in many applications like autonomous driving and video surveillance. Recently, deep fully convolution network (FCN) based methods \cite{Chen2017deeplab, Chen2018deeplabv3plus} have achieved remarkable results in semantic segmentation. Extensive methods have been investigated to improve the performance by introducing sophisticated models with a large number of parameters. 
To preserve the detailed semantic structures in the dense estimation, many state-of-the-art FCN based methods \cite{Zhang2018exfuse, Chen2017deeplab, Chen2017deeplabv3, Chen2018deeplabv3plus} maintain a series of high-resolution feature maps by applying a small overall stride \cite{Chen2018deeplabv3plus}, which causes heavy computations and limits the practicability of semantic segmentation. 
For example, Chen proposed DeepLabV3+ \cite{Chen2018deeplabv3plus} and achieved state-of-the-art performance on many open benchmarks such as Pascal VOC \cite{Everingham2014pascal} and Cityscapes \cite{Cordts2016Cityscapes}. However, this is obtained back-boned on a large model: Xception-65 \cite{Chollet2017xception}, which contains more than 41.0M parameters and 1857G FLOPS and runs at 1.3 FPS on a single 1080Ti GPU card if the output stride is set to 16. Even worse, 6110G FLOPS will be needed and running at 0.4 FPS with output stride of 8. Similar situation can be found in lightweight models (see Fig.~\ref{fig:flops}).
\par

\par
One instant way to address this limitation is to reduce the resolution of a series of feature maps via sub-sampling operations like pooling and convolution striding. However, unsatisfactory estimation  accuracy will be incurred for the huge loss of detailed information.

\par
How to solve the dilemma and find a better trade-off between the accuracy and efficiency have been discussed for a long time. Knowledge distillation (KD), introduced by Hinton \cite{Hinton2015distill} to the field of deep learning, has attracted much attention for its simplicity and efficiency. The knowledge in \cite{Hinton2015distill} is defined as soft label output from a large teacher network, which contains more useful information, such as intra-class similarity, than one-hot encoding. The student network is supervised by both soft labels and hard one-hot labels simultaneously, reconciled by a hyper-parameter to adjust the loss weight. Following KD \cite{Hinton2015distill}, many methods \cite{Romero2015fitnets,Huang2017like, Zagoruyko2017atnet, Yim2017agift, Jangho2018paraphrasing} are proposed to regulate the intermediate features. 
However, these methods are mainly designed for the image-level classification task without considering the spatial context structures. Moreover, in the semantic segmentation task, the feature maps from the teacher and student usually have inconsistent context and mismatched features. Thus these methods are improper to be used for semantic segmentation directly.

\par
In this paper, we propose a new knowledge distillation method tailored for semantic segmentation. We aim to learn efficient compact FCNs (\ie, student) by distilling the rich and powerful knowledge from the accurate but heavy teachers with larger overall stride. 
Firstly, unlike other methods that force the student to mimic the output values from the teacher network directly, we rephrase the rich semantic knowledge from the teacher into a compact representation. The student is trained to match this implicit information. The knowledge translating is achieved relying on an auto-encoder pre-trained on the teacher features in an unsupervised manner, which reformulates the knowledge from the teacher to a compact format that is more easier to be comprehended by the student network.
The behind intuitions are quite straightforward: 
Directly transferring the outputs from teacher overlooks the inherent differences of network architecture between two models. Compact representation, on the other hand, can help the student focus on the most critical part by removing redundancy knowledge and noisy information.
Furthermore, we also propose an affinity distillation module to regulate relationships among widely separated spatial regions between teacher and student. Compared to large models, small models with fewer parameters are hard to capture long-term dependencies and can be statistically brittle, due to the limited receptive field. The proposed affinity module alleviate the situation by explicitly computing pair-wise non-local interactions cross the whole image.

To summarize, our main contributions are as follows.
\begin{itemize}
	\itemsep -2pt
	\item We propose a new knowledge distillation method tailored for semantic segmentation that reinterprets
	the output from the teacher network to a re-represented latent domain, which is easier to be learned by the compact student model.
	
	\item We come up with an affinity distillation module to help the student network capture long-term dependencies from the teacher network. 
	
	\item We validate the effectiveness of methods under various settings.  (1) Our method improves the performance of the student model by a large margin (\%2) without introducing extra parameters or computations.  (2) Our model achieves at least comparable or even better results with only 8\% FLOPS compared to the model with large resolution outputs.
	
\end{itemize}

%% file: relatedwork.tex
\section{Related Work}
In this section, we review the literatures that are most relevant to our work, including the state-of-the-art researches for semantic segmentation and related methods for knowledge distillation.

\textbf{Semantic Segmentation}
Semantic segmentation is a fundamental and challenging task in computer vision. With the advent of fully convolution networks, lots of progress have been made. Among all the factors for these successes, rich spatial information and sizable receptive fields are two important clues \cite{Chen2017deeplabv3, Chen2017deeplab, Chen2018deeplabv3plus}.

Chen \etal proposed DeepLab-CRF \cite{Chen2017deeplab}, which applies a dense CRF as a post processing step to refine the segmentation results and capture better boundaries on the top of CNN. 
This method is extended by CRF-RNN \cite{Zheng2015crfasrnn}, in which CRF is implemented as an inner layer embedded in a network for end-to-end learning. Lin \etal \cite{Lin2017refinenet} proposed a multi-path RefineNet to output high-resolution results, by exploiting long-range residual modules to capture all information when down sample operations is performed. 
Recently, Chen \etal proposed DeepLabV3 \cite{Chen2017deeplabv3} and DeepLabV3+ \cite{Chen2018deeplabv3plus} that applied atrous convolution operation to effectively enlarge the reception field and capture rich semantic information. 
These methods improve the performance by outputting high resolution feature maps to alleviate the loss of details and boundaries. 
However, considering the limit of GPU resources and computational efficiency, $\frac{1}{8}$ or even more denser $\frac{1}{4}$ size of inputs resolution are not realistic in current model design. For example, when ResNet-101 \cite{He2015residual} uses the atrous convolution to output 16 times smaller feature maps, much more computation and storages will be used in the last 9 convolution layers. Even worse, 26 residual blocks (78 layers!) will be affected if the output features that are 8 times smaller than the input are desired. In this paper, we propose a novel method that successfully compresses these dense information from the teacher network and distill the compact knowledge to the student network with a low-resolution output. 

\begin{figure*}
	\centering
	\includegraphics[width=0.708\textwidth]{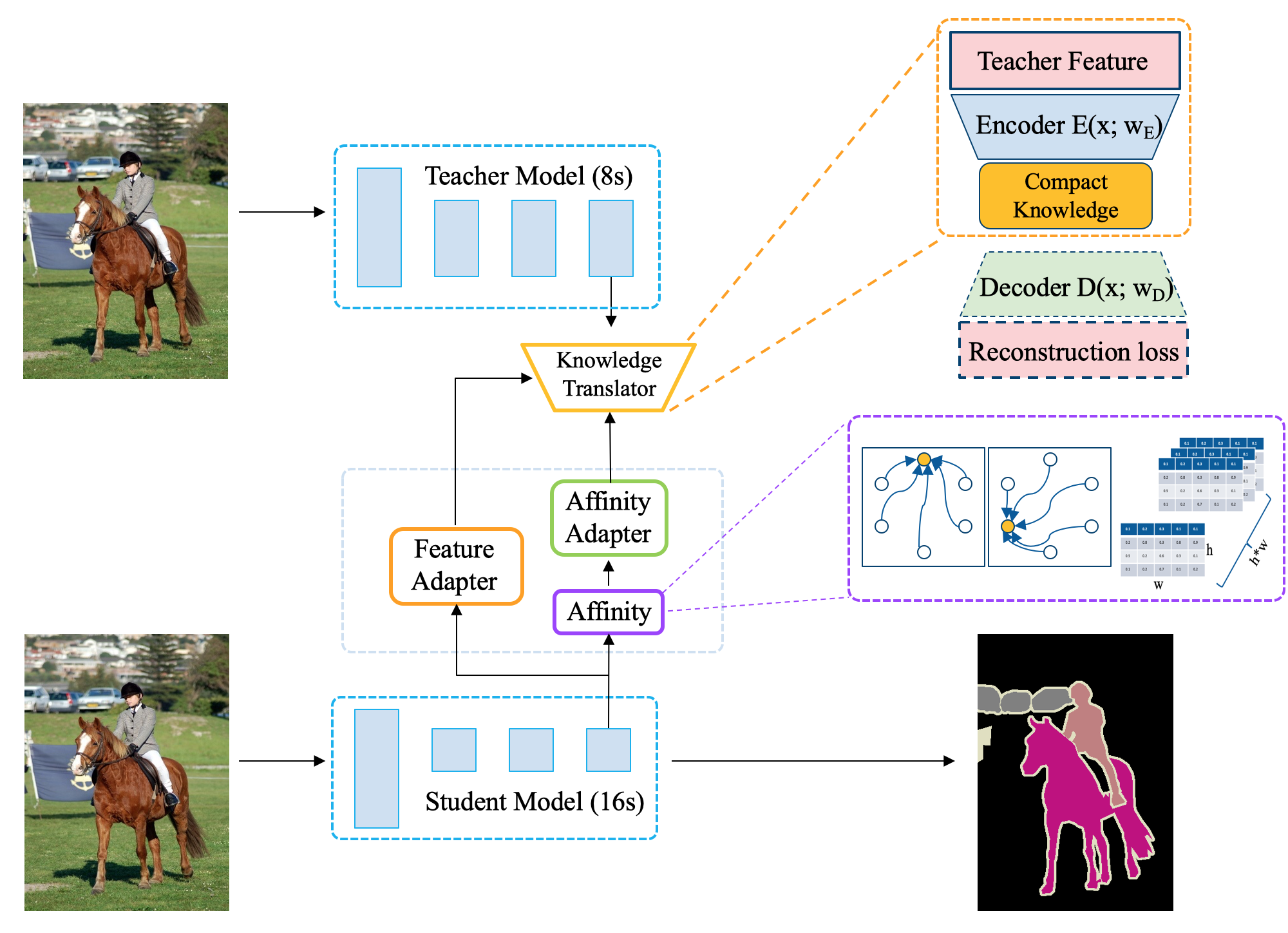} 
	\caption{The detailed framework of our knowledge adaptation method tailored for semantic segmentation. The teacher network is frozen and outputs high resolution feature maps. The student network outputs small size of feature maps and is updated by both ground truth labels and the knowledge defined in a compressed space and affinity information.}
	\label{fig:main}
\end{figure*}

\textbf{Knowledge Distillation}
The research of \cite{Hinton2015distill} is the pioneering work that exploits knowledge distillation for the image classification task. The knowledge is defined as the soft output from the teacher network which provides much more useful information, such as intra-class similarity and inner-class diversity, than one-hot encoding. The soften degree is controlled by a hyper-parameter temperature, $T$. 
The student network is supervised by two losses reconciled by a loss weight. Despite of its effectiveness on image classification, there are some limitations for its application in the semantic segmentation task: (1) Authors in \cite{Romero2015fitnets} tried to force the student to mimic the output distribution of a teacher network in the decision space, where useful context information are cascaded. (2) The knowledge required for image-level classification are similar between two models, because both models capture global information. But the decision space may different for semantic segmentation, because two models have different abilities to capture context and long range dependencies, depending on the network architecture. (3) The hyper-parameter temperature is sensitive to tasks and is hard to tune, especially on large benchmarks.

Following \cite{Hinton2015distill}, many other methods are proposed for knowledge distillation. Romero et al.\ proposed FitNet \cite{Romero2015fitnets}, for the purpose of learning intermediate representation by directly aligning feature maps, which may not be a good choice for overlooking the inherent differences between two models, such as spatial resolution, channel numbers, and network architecture. Meanwhile, significantly different abstracting capability between two models may make this situation severe.
Attention transfer \cite{Zagoruyko2017atnet} (AT) aims to mimic the attention map between student and teacher models. It is based on the assumption that the summation of feature maps across channel dimension can represent attention distribution in the image classification task. However, this assumption may not suit pixel-wise segmentation task, because different channels are representing activations of different classes and simply summing up across channels will end up with mixed attention maps. In our work, we propose a new affinity distill module to transfer these long-rage dependencies among widely separated spatial regions from a teacher model to a student model.

%% file: methods.tex
\newcommand\mynorm[1]{\left\lVert#1\right\rVert}

\section{Proposed Method}

With the help of the atrous convolution operation, a network with a small overall output stride often outperforms the one with a large overall output stride for capturing detailed informations, as shown in Figure \ref{fig:flops}. Inspired by this, we propose a novel knowledge distillation method tailored for semantic segmentation. As shown in Figure \ref{fig:main}, the whole framework involves two separate networks: one is the teacher network, which outputs features with larger resolution (e.g., 8s overall stride), the other is the student network, which has smaller outputs (e.g., 16s overall stride) for fast inference. 
The knowledge is defined as two parts: (1) The first part is designed for translating the knowledge from the teacher network to a compressed space that is more informative. The translator is achieved by training an auto-encoder to compress the knowledge to a compact format that is easier to be learned by the student network, otherwise much harder due to the inherent structure differences. (2) The second part is designed to capture long-range dependencies from the teacher network, which is difficult to be learned for small models due to the limited receptive field and abstracting capability.
More details are provided in the following sections.

\subsection{Knowledge Translation and Adaptation}
Benefiting from the atrous convolution operation, FCNs can maintain detailed information while capturing a large receptive field. Although the performance is improved, large computation overheads are introduced and will grow exponentially as the output stride becomes smaller, as shown in Figure \ref{fig:flops}. 
In this section, we propose to utilize a large teacher model with high feature resolution to teach a lightweight student network with low feature resolution.

An auto-encoder, which tries to reconstruct the input, is capable of capturing useful and important information. We train an auto-encoder for mining the implicit structure information and translating the knowledge to a format that is easier to be comprehended and replicated by the student network. Compared with low-level and middle-level features, which are either general across different models or challenging to be transferred due to the inherent network differences, high-level features are more suitable for our situation. 
In our method, the auto-encoder takes the last convolution features from the teacher model as input and is composed of three strided convolution layers and symmetrical deconvolution layers. Suppose that we have two networks, namely, the student network $S$ and the teacher network $T$ and the last feature maps of the two models are $\Phi_{s}$ and $\Phi_{t}$, respectively. The training process is completed by using a reconstruction loss in Eq. \eqref{eq:reconstruction_loss},
\begin{figure}
	\centering
	\includegraphics[width=0.465\textwidth,height=.45\textwidth]{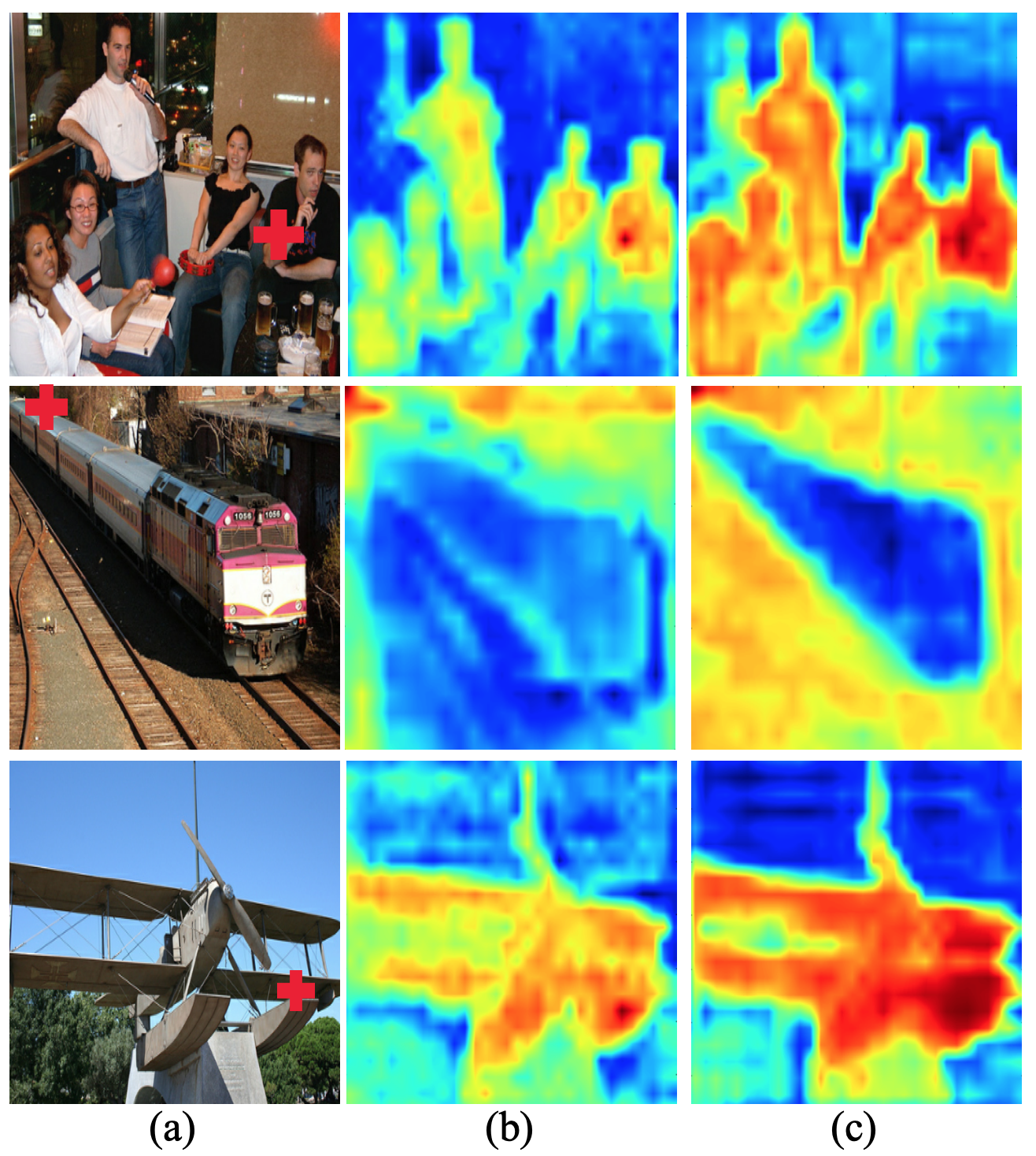} 
	\caption{The effect of affinity distillation module (better visualized in color). (a) input image and random selected point with red '+'. (b) affinity map of the given point of student model without affinity distillation module. (c) affinity map enhanced by our affinity distillation module.}
	\label{fig:adm}
\end{figure}

\begin{equation}
\label{eq:reconstruction_loss}
L_{\text{ae}}  = \| \Phi_{t} - \mathit{D}(\mathit{E}( \Phi_{t}))  \|^{2} + \alpha\|\mathit{E}( \Phi_{t}) \|_1
\end{equation}
where $\mathit{E}(\cdot)$ and $\mathit{D}(\cdot)$ represent encoder and decoder, respectively. One common issue in training the auto-encoder is that the model may learn little more than an identity function, implying the extracted structure knowledge is more likely to share the same pattern with the input features. As the $l1$ norm is known to produce sparse representations, similar strategy \cite{Ayinde2018deep} is utilized by regularizing both weights and the re-represented space. The weight for regularization loss $\alpha$ is set to $10^{-7}$ for all experiments.
In order to solve the problem of feature mismatching and decrease the effect of the inherent network difference of two models, the feature adapter is utilized by adding a convolution layer.

\par
Relying on the pre-trained auto-encoder, the transferring process is formalized in Eq. \eqref{eq:transfer_loss}, 
\begin{equation}
\begin{aligned}
\label{eq:transfer_loss}
L_{\text{adapt}} 
= \frac{1}{|I|}\sum_{j\in I}
\mynorm
{
	\frac{ C_{f}(\Phi_{s}^{j}) }{ \| C_{f}(\Phi_{s}^{j}) \|_{q}} - \frac{E(\Phi_{t}^{j})}{\|E(\Phi_{t}^{j}) \|_{q}} 
}_p
\end{aligned}
\end{equation}
where $\mathit{E}$ represents the pre-trained auto-encoder. $I$ denotes the indices of all student-teacher pairs in all positions. $\mathit{C_{f}}$ is the adapter for the student features, which uses a 3 $\times$ 3 kernel with stride of 1, padding of 1, BN layer and ReLU activation function. The features are normalized before matching.
$p$ and $q$ are different normalization types to normalize the knowledge for stability. 

\subsection{Affinity Distillation Module}
Capturing long-range dependency is important and can benefit the task of semantic segmentation. As described in \cite{Wang2018nonlocal}, it is easier to be captured by deep stacked convolution layers with large receptive field. Small networks, on the other hand, have limited ability to learn this knowledge due to the deficient abstracting capability. We propose a new affinity distillation module by explicitly extracting long-range, non-local dependencies from the big teacher model. Details are described below.

In the case of studying, sometimes it would be more efficient to learn new knowledge by providing extra difference or affinity information. Inspired by this, we define the affinity in the network by directly computing interactions between any two positions, regardless of their spatial distances. As a result, the pixels with different labels will generate a low response and a high response for the pixels with the same labels. Let feature maps of the last layer to be $\Phi$ with size of $h\times w\times c$, where $h$, $w$ and $c$ represent the number of height, width, and channels, respectively. 
The affinity matrix $A \in \mathbb{R}^{m*m}$ can be calculated by Eq. \eqref{eq:affinity_matrix}, where $m$ equals to $h$$\times$$w$, $i$ and $j$ are the indexes for vectorized $\Phi$: 
\begin{equation}
\label{eq:affinity_matrix}
A(\Phi)_{i,j} = \frac{1}{h\times w} \cdot  \frac{\Phi_{i}}{\|\Phi_{i} \|_2} \cdot  \frac{\Phi_{j}}{\|\Phi_{j} \|_2},
\end{equation}
where $A(\Phi)$ denotes the affinity matrix corresponding to the feature map $\Phi$ with spectral normalization.

We use $\ell_2$ loss to match affinity matrix between teacher and student models, which is defined as Eq. \eqref{eq:affinity_loss}

\begin{equation}
\label{eq:affinity_loss}
L_{\text{aff}} = \sum_{i} \| (A_s(C_{a}(\Phi_{s}))) - A_t(\mathit{E}(\Phi_{t})) \|_2
\end{equation}
where $\mathit{E}(\Phi_{t})$ is the translated knowledge from teacher, $C_{a}$ is the adapter for student affinity and $i$ is the location index of the feature map. 

To visualize the effect of the affinity distillation module, some examples are presented in Figure \ref{fig:adm}. Given one random selected point, the response between this point and all other separated spatial regions are shown in (b) and (c). As can be seen, the student network fails to capture this long-range dependency and only local similar patterns are highlighted. With the help of our method, long-range or even global information are captured and can be used to make more robust decision.

\subsection{Training Process }

Our proposed method involves a teacher net and a student net. As is presented in Algorithm \ref{alg:trainingprocess}, the teacher net is pre-trained and the parameters are kept frozen during the training the transferring process. The student net is supervised by three losses: cross entropy loss $L_{\text{ce}}$ with ground truth label, adaptation loss $L_{\text{adapt}}$ in Eq. \eqref{eq:transfer_loss} and affinity transferring loss $L_{\text{aff}}$ in Eq. \eqref{eq:affinity_loss}. Three losses are reconciled by the loss weights of $\beta$ and $\gamma$, which are set to 50 and 1 respectively in all our experiments. $\textit{W}_{E}$, $\textit{W}_{D}$ and $\textit{W}_{S}$ denote parameters for encoder, decoder and student model, respectively.

\begin{algorithm}
	
	\caption{Training Process of Our Method}  
	\begin{algorithmic}
		\REQUIRE ~~ Already trained teacher network $T$.\\
		\STATE { \textbf{STAGE 1}: Training auto-encoder for teacher network. }\\
		{\textrm{INPUTS}: Knowledge from teacher network $\Phi_{t; W_t}$} \\
		$\qquad$ {$\textit{W}_{E} = \mathop{\arg\min}_{\textit{W}_{E}, \textit{W}_{D}} L_{\text{ae}}$ ($\Phi_{t; W_t}$)}

		\STATE { \textbf{STAGE 2}: Training student network. }\\
		{\textrm{INPUTS:}  Encoder Parameters $\textit{W}_{E}$} \\
		$\qquad$ {$\textit{W}_{S} = \mathop{\arg\min}_{W_{S}}  L_{\text{ce}} + \beta L_{\text{adapt}} + \gamma L_{\text{aff}}$}
		
	\end{algorithmic}
	\label{alg:trainingprocess}
\end{algorithm}

%% file: experiment.tex
\section{Experiments}

\begin{table}[!htb]
	\small
	\begin{center}
		\caption{Ablations for the proposed method. \textbf{T}: Teacher model has a output stride of 8s. \textbf{S}: Student model (following the implementation of \cite{Sandlermobilenetv22018}, without ASPP and decoder) has a output stride of 16s. \textbf{KA} represents knowledge adaption. The FLOPS is estimated with input size of 513$\times$513. For fair comparison, all the models are trained on the Pascal VOC \textit{trainaug} set \cite{Hariharan2011semantic} tested on the \textit{val} set without pre-training on the COCO dataset. As can be seen, our proposed method with small feature resolution outperforms the student model with large feature resolution by only 31\% FLOPS.}
		\label{tab:component}
		\scalebox{.92}{
		\begin{tabular}{lc c c}
			\toprule
			Method & mIOU\%) & FLOPS &Params\\
			\noalign{\smallskip}
			\midrule
			\noalign{\smallskip}
			T: ResNet-50-8s \cite{He2015residual} & 76.21 & 90.24B &26.82M\\
			S1: MobileNetV2-16s \cite{Sandlermobilenetv22018} & 70.57 &5.50B &2.11M \\
			S2: MobileNetV2-8s \cite{Sandlermobilenetv22018} & 71.90 &17.70B &2.11M \\
			\noalign{\smallskip}
			\midrule
			\noalign{\smallskip}
			S1+affinit-16s & 71.53 &5.5B &2.11M\\
			S1+KA+affinity-16s & 72.50 &5.5B & 2.11M\\
			\bottomrule
		\end{tabular}
	}
	\end{center}
	\vspace{-0.3cm}
\end{table}

In this section, we first introduce the datasets and the implementation details of our experiments. Extensive ablation studies are followed to investigate the effectiveness of our proposed methods. Finally, we report our results and make a comparison with other lightweight models on three popular benchmarks: Pascal VOC \cite{Everingham2014pascal}, Cityscapes \cite{Cordts2016Cityscapes} and Pascal Context \cite{Mottaghi2014}.

\subsection{Datasets}
\textit{Pascal VOC.} This dataset contains 1,464 images for training, 1,449 for validation, and 1,456 for testing. It contains 20 foreground objects classes and an extra background class. In addition, the dataset is augmented by extra coarse labeling provided by \cite{Hariharan2011semantic}. The final performance is measured in terms of pixel intersection-over-union (mIOU) averaged across the 21 classes.

\textit{Cityscapes.} This dataset focuses on semantic understanding of urban street scenes, which contains high-resolution images with 1024$\times$2048 pixels and sense pixel-wise annotations.
The dataset includes 5,000 finely annotated images collected from 50 cities, and is split with 2,975 for training, 500 for validation, and 1,525 for testing. Following the evaluation protocol, 19 output of 30 semantic labels are used for evaluation.

\textit{Pascal Context.} The dataset contains 10,103 images in total, out of which 4,998 are used for training and 5,105 are used for validation. Following \cite{Mottaghi2014}, methods are evaluated on the most frequent 59 classes with one background class.

\subsection{Implementation Details}
The MobileNetV2, recently proposed by Sandler \etal \cite{Sandlermobilenetv22018}, has attracted much attention for its computation efficiency and optimal trade-offs between accuracy and the number of operations measured by FLOPS, actual latency, and the number of parameters.
There are also MobileNetV2-1.3 and MobileNetV2-1.4, which are model variants with a width multiplier of 1.3 and 1.4, respectively.
The mobile segmentation models in \cite{Sandlermobilenetv22018} use a reduced form of DeepLabV3 \cite{Chen2017deeplabv3}. Built on this strong baseline, our method significantly boosts the performance without introducing extra parameters and computation overheads.

\textbf{Training teacher network.} To demonstrate the effectiveness of our method, we select two totally different teacher models, ResNet-50 \cite{He2015residual} and Xception-41 \cite{Chollet2017xception}. Both atrous convolution and atrous spatial pyramid pooling (ASPP) are utilized to obtain a series of feature maps with large size.
\begin{table}[!t]
	\small
	\begin{center}
		\caption{The performance on the Pascal VOC 2012 \textit{val} data set with different student and teacher networks.  MobilNetV2 is tailored with a width-multiplier. Performances are obtained by training on \textit{trainaug} set.}
		\label{tab:different_nets}
		\scalebox{.92}{
		\begin{tabular}{lc c c}
			\toprule
			Method & mIOU(\%) & FLOPS &Params\\
			\noalign{\smallskip}
			\midrule
			\noalign{\smallskip}
			T1: ResNet-50 \cite{He2015residual} & 76.21 &90.24B &26.82M\\
			T2: Xception-41 \cite{Chollet2017xception} & 77.2 & 74.69B &27.95 \\
			S1: MobileNetV2-1.0 \cite{Sandlermobilenetv22018} & 70.57 &5.50B &2.11M \\
			S2: MobileNetV2-1.3 \cite{Sandlermobilenetv22018} & 72.60 &9.02B &3.38M \\
			S3: MobileNetV2-1.4 \cite{Sandlermobilenetv22018} & 73.36 &10.29B &3.88M \\
			\noalign{\smallskip}
			\midrule
			\noalign{\smallskip}
			T1+S1+our method & 72.50 &5.5B &2.11M\\
			T2+S1+our method & 72.40 &5.5B &2.11M\\
			T1+S2+our method & 74.26 &28.72B & 3.38M\\
			T1+S3+our method & 74.07 &32.91B & 3.88M\\
			\bottomrule
		\end{tabular}
	}
	\end{center}
\end{table}

\begin{figure*}
	\begin{center}
		\includegraphics[width=17cm]{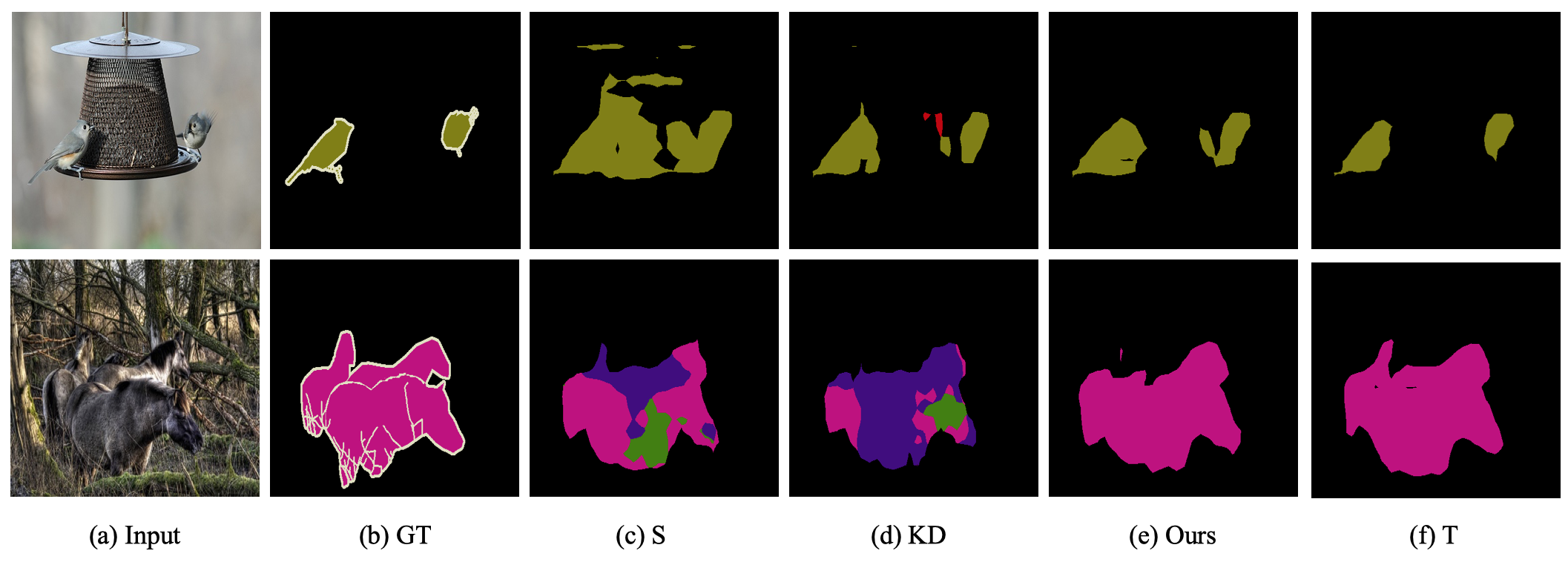}
	\end{center}
	\caption{Comparison of segmentation results. (a) Input image. (b) Ground truth. (c) The results of the student network, MobileNetV2 \cite{Sandlermobilenetv22018}. (d) Results of the knowledge distillation \cite{Hinton2015distill} with MobileNetV2 \cite{Sandlermobilenetv22018}. (e) Results of our proposed method with MobileNetV2 \cite{Sandlermobilenetv22018}. (f) Results of the teacher network, which is ResNet50 \cite{He2015residual}.}
	\label{fig:results}
\end{figure*}
We use mini-batch stochastic gradient descent (SGD) with batch size 16 (at least 12), momentum 0.9, and weight decay $4\times10^{-5}$ in training. Similar to \cite{Chen2018deeplabv3plus}, we apply the “poly” learning rate strategy with power 0.9. The initial learning rate is 0.007. General data augmentation methods are also used in network training, such as randomly flipping the images and randomly performing scale jitter.
For the Pascal VOC dataset, the training process can be split into two steps. First, we train 300K iterations on the COCO dataset, then 30K iterations on the \textit{trainaug} dataset \cite{Hariharan2011semantic}.
For the Cityscapes dataset, we do not pre-train our model on the COCO dataset for fair comparison. We train 90K iterations on the \textit{train-fine} dataset, which is fine tuned on \textit{trainval} and \textit{train-coarse} to evaluate on \textit{test} dataset.
For the Pascal Context dataset, the COCO dataset is not used for pre-training. 30k iterations are trained on the \textit{train} set and evaluated on the \textit{val} set.

\textbf{Training auto encoder.}  We finished the auto-encoder training within one epoch with a learning rate of $0.1$.  Large weight decay of $10^{-4}$ is used to attribute low energy to a smaller portion of the input points.

\textbf{Training the whole system.} Most of the training parameters are similar to the process of training the teacher network, except that our student network does not involve the ASPP and the decoder, which are exactly the same with \cite{Sandlermobilenetv22018}. With the help of atrous convolution, low resolution feature maps are generated. During the training process, the parameters of the teacher net $\mathit{W}_T$ and the parameters for auto-encoder $\mathit{W}_{E}$ are fixed without updating.

\begin{figure}
	\centering
	\includegraphics[width=8cm]{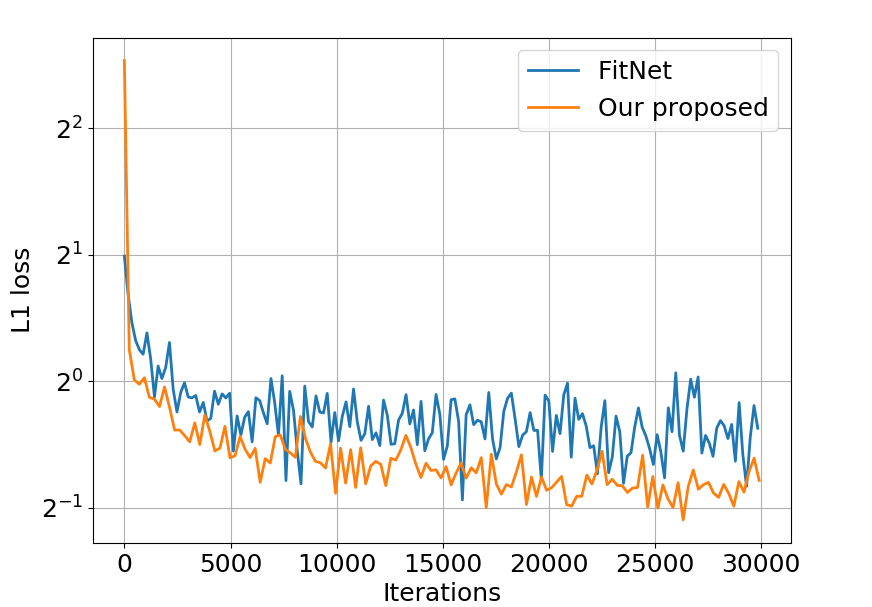}
	\caption{The L1 loss curve for knowledge transferring process. Our method using translator and adapter makes it easier for student network to learn and replicate the knowledge. }
	\label{fig:losses}
\end{figure}

\subsection{Ablation Study}
In this section, we describe the effectiveness of every component of our proposed method with different settings.

\begin{table}[!b]
	\centering
	\small
	\caption{The performance on the Pascal VOC 2012 \textit{val} set in comparison with KD \cite{Hinton2015distill} and FitNet \cite{Romero2015fitnets}. All the results are achieved by training only on the Pascal VOC \textit{trainaug} set. }
	\label{tab:kd}
	\begin{tabular}{lc }
		\toprule
		Method & mIOU(\%) \\
		\noalign{\smallskip}
		\midrule
		\noalign{\smallskip}
		T: ResNet-50 \cite{He2015residual} & 76.21 \\
		S: MobileNetV2 \cite{Sandlermobilenetv22018} & 70.57 \\
		\noalign{\smallskip}
		\midrule
		\noalign{\smallskip}
		S+KD \cite{Hinton2015distill} ($t=2$) & 71.32\\
		S+KD \cite{Hinton2015distill} ($t=4$) & 71.21\\
		S+KD \cite{Hinton2015distill} ($t=8$) & 70.74\\
		S+FitNet \cite{Sandlermobilenetv22018} & 71.30\\
		S+Ours& 72.50\\
		\bottomrule
	\end{tabular}
	\vspace{-0.3cm}
\end{table}

\subsubsection{Ablation for the knowledge adaption and the affinity distillation module.}
In order to make use of rich spatial information, we propose to translate the knowledge from the teacher and force the student to mimic this compact format. The affinity distillation module is also proposed to make up the limited receptive field of the small student model. To show a better understanding, we visualize the effect of the affinity distillation module in Figure \ref{fig:adm}. It can be seen from Figure \ref{fig:adm}, that more context and long-range dependencies are captured with the help of our proposed method. We show the statistic results in Table \ref{tab:component}, where performance is evaluated using mIOU. The model is tested in one single scale on the Pascal VOC \textit{val} set without pretraining on the COCO dataset. As can be seen, the affinity distillation module boosts the performance from 70.57 to 71.53, and another 0.97 point with the help of knowledge adaption. Because the affinity matrix mismatches if two models have different output features, in order to show the effect of a single affinity module, we resize the feature maps to the same dimension.
Our MobileNetV2 with output stride of 16 even outperforms MobileNetV2 with output stride of 8, using only 31\% FLOPS. More comparisons with different output stride settings can be found in Figure \ref{fig:main}, where our 16s model performs even better than the baseline model with 4s output by using only 8\% FLOPS without introducing extra parameters.

\begin{table}[!b]
	\small
	\begin{center}
		\caption{Comparison with other lightweight models on the Pascal Context \textit{val} set. ``-'' means not provided.}
		\label{tab:pascal_context}
		\begin{tabular}{lc c c c}
			\toprule
			Method   &FLOPS &Params &mIOU(\%)  \\
			\noalign{\smallskip}
			\midrule
			\noalign{\smallskip}
			FCN-8s \cite{Long2015fully} &135.21G &1.48M &37.8 \\
			ParseNet \cite{Liu2015semantic}  &162.82G &21.53M &40.4 \\
			Piecewise\_CRF \cite{Lin2015efficient} &\textgreater100G & - & 43.3 \\
			DAG\_RNN \cite{Shuai2017dag} &\textgreater100G & - & 42.6 \\
			\hline \\
			MobileNetV2 \cite{Sandlermobilenetv22018}  &5.52G &2.12M& 39.9 \\
			Ours  &5.52G &2.12M & 41.2 \\

			\bottomrule
		\end{tabular}
	\end{center}
	\vspace{-0.3cm}
\end{table}

\subsubsection{Ablation for different networks.}
From \cite{Sandlermobilenetv22018}, MobileNetV2 tailors the framework to achieve different accuracies, by using width-multiplier as a tunable hyper-parameter, which is used to adjust the trade-off between accuracy and efficiency. In our experiments, we choose width-multipliers of 1.3 and 1.4, which are implemented with official pre-trained models on ImageNet. In order to validate the effectiveness of our proposed method, we choose two totally different network architectures, ResNet-50 \cite{He2015residual} and Xception-41 \cite{Chollet2017xception}. The results are shown in Table \ref{tab:different_nets}. The performance of MobileNetV2-1.0 gains $1.93$ and $1.83$ improvements under the guidance of ResNet-50 and Xception-41, respectively. Improvements of $1.66$ and $0.71$ are also observed with different student networks: MobileNetV2-1.3 and MobileNetV2-1.4.

\begin{table}[!b]
	\small
	\centering
	\caption{Comparison with other lightweight models on the Pascal VOC 2012 \textit{val} data set. Speed is tested on single 1080Ti GPU with input size of 513 $\times$ 513. The baseline is our implementation of MobileNetV2. }
	\label{tab:vocval}
	\scalebox{0.884}{
	\begin{tabular}{lc c c c}
		\toprule
		Method & basemodel &FPS &mIOU(\%)  \\
		\noalign{\smallskip}
		\midrule
		\noalign{\smallskip}
		CRF-RNN \cite{Zheng2015crfasrnn} &VGG-16 \cite{Simonyan2014vggnet} & 7.6 &72.9 \\
		MultiScale \cite{Yu2015multiscale} &VGG-16 \cite{Simonyan2014vggnet} & 16.7 &73.9 \\
		DeeplabV2 \cite{Chen2017deeplab} &VGG-16 \cite{Simonyan2014vggnet} &16.7 &75.2 \\
		MobileNetV2 \cite{Sandlermobilenetv22018} &MobileNet &120.7 & 75.3 \\
		Baseline  &MobileNet  &120.7  &74.8 \\
		Ours &MobileNet &120.7 & 75.8 \\
		\bottomrule
	\end{tabular}
}
\end{table}

\begin{figure*}
	\begin{center}
		\includegraphics[width=.808\textwidth]{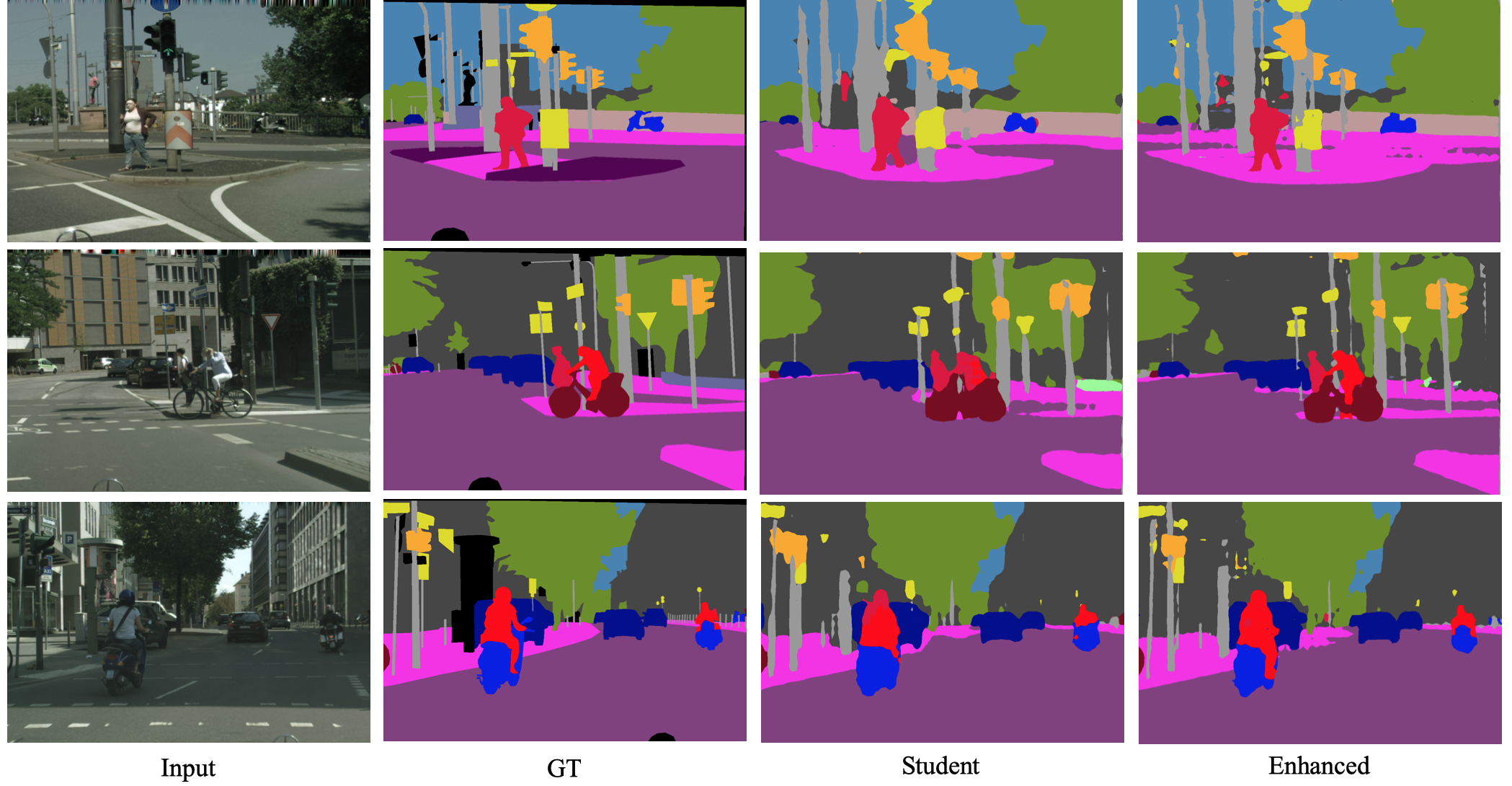}
	\end{center}
	\vspace{-0.1cm}
	\caption{Example results on the Cityscapes dataset. From left to right are: (1) Input images, (2) Ground truth, (3) The results of the student net (4) The results of our proposed method.}
	\label{fig:cityscapes}
\end{figure*}

\begin{table}[!b]
	\small
	\begin{center}
		\caption{Performance and computation comparisons of our proposed method against other light-weight models on the Cityscapes \textit{val} and \textit{test} data sets. The running times are all computed with input size of 1025 $\times$ 2049. ``-'' means not provided.}
		\label{tab:city-speed-comp}
		\begin{tabular}{ccccccc}
			\toprule
			\multicolumn{2}{l}{\multirow{2}*{Method}} & \multicolumn{1}{l}{\multirow{2}*{Year}} & \multicolumn{2}{l}{\multirow{2}*{Time}} & \multicolumn{2}{c}{mIOU (\%)} \\
			\cmidrule(lr){6-7}
			\multicolumn{2}{c}{}& \multicolumn{1}{c}{} & \multicolumn{2}{c}{} & \emph{val} & \emph{test} \\
			\noalign{\smallskip}
			\hline
			\noalign{\smallskip}

			\multicolumn{2}{l}{DeepLabV2 \cite{Chen2017deeplab}} &'16 & \multicolumn{2}{l}{652.9ms} & - & 71.4 \\
			\multicolumn{2}{l}{Dilation-10 \cite{Yu2015multiscale}} &'17 & \multicolumn{2}{l}{3549.5ms} & - & 67.1 \\
			\multicolumn{2}{l}{PSPNet \cite{Zhao2017psp}} &'17 & \multicolumn{2}{l}{2647.4ms} & - & 80.2 \\
			\multicolumn{2}{l}{ResNet38 \cite{Wu2017resnet38}} &'17 & \multicolumn{2}{l}{3089.9ms} & 77.86 & 78.4 \\
			\noalign{\smallskip}
			\hline
			\noalign{\smallskip}
			\multicolumn{2}{l}{SegNet \cite{Badrinarayanan2017segnet}} &'15 & \multicolumn{2}{l}{89.2ms} & - & 57.0 \\
			\multicolumn{2}{l}{ENet \cite{Paszke2016enet}} &'16 & \multicolumn{2}{l}{19.3ms} & - & 58.3 \\
			\multicolumn{2}{l}{SQ \cite{Treml2016sq}} &'16 & \multicolumn{2}{l}{-} & - & 59.8 \\
			\multicolumn{2}{l}{ICNet \cite{Zhao2017icnet}} &'18 & \multicolumn{2}{l}{33.0ms} & 67.7 & 70.6 \\
			\noalign{\smallskip}
			\hline
			\noalign{\smallskip}
			\multicolumn{2}{l}{MobilenetV2 \cite{Sandlermobilenetv22018}} &'18 & \multicolumn{2}{l}{38.0ms} & 68.9 & 70.2 \\

			\multicolumn{2}{l}{Ours} &- & \multicolumn{2}{l}{38.0ms} & 71.0 & 72.7 \\

			\bottomrule
		\end{tabular}
	\end{center}
	\vspace{-0.3cm}
\end{table}

\subsubsection{Ablation for other method for knowledge distillation.}
In this experiment, we make comparisons with other knowledge distillation methods: KD \cite{Hinton2015distill} and FitNet \cite{Romero2015fitnets}, which are designed for image-level classification. The knowledge defined in \cite{Hinton2015distill} is the soft label output by a teacher network. The soften degree is controlled by a hyper-parameter temperature $t$, which has a significant influence on the distillation and learning processes. We set $t$ to 2, 4, 6.
To make fair comparisons, we bilinearly upsample the logits map to the size of the teacher network. The results are evaluated on the Pascal VOC \textit{val} dataset. All results are achieved without pre-training on COCO dataset.
FitNet \cite{Romero2015fitnets}, different from KD, tries to match the intermediate representation between two models. But this requires similar network design. In our experiments, we directly upsample the feature map of the last layer and add a $\ell_2$ loss. The loss curve is shown in Figure \ref{fig:losses}. Our proposed method successfully translate the knowledge from teacher to a format that is easier to be learned.
As shown in Table \ref{tab:kd}, fluctuation of mIOU is observed with different settings of $T$. Our method achieves better performances than KD, with all the hyper-parameters fixed across all experiments and datasets. Our method also outperforms FitNet by $1.2$ points, indicating that the knowledge defined by our method alleviates the inherent difference of two networks.
Compared with the traditional methods, the qualitative segmentation results in Figure \ref{fig:results} visually demonstrate the effectiveness of our distillation method for objects that require more context information, which is captured by our proposed affinity transfer module. On the other hand, the knowledge translator and adapter reduce the loss of the detailed information and produce more consistent and detail-preserving predictions, as shown in Figure \ref{fig:cityscapes}.

\subsubsection{Comparing with other lightweight  models.}
We first test our method on the Pascal Context dataset. The results are shown in Table \ref{tab:pascal_context}. Our proposed method boosts the baseline by 1.3 points.

Then we compare our proposed method with other state-of-the-art light weight models on the Pascal VOC \textit{val} dataset. The results are shown in Table \ref{tab:vocval}. Our model yields mIOU 75.8, which is quantitatively better than several methods that do not care about speed. It also improves the baseline of MobileNetV2 by about 1 point.

Finally, we testify the effectiveness of our method on the Cityscapes dataset. It achieves 70.3 and 72.7 mIOU on the \textit{val} and \textit{test} data sets, respectively. Even built on a highly competitive baseline, our method boosts the performance by 2.1 and 2.5 points, without introducing extra parameters and computations overheads, as shown in Table \ref{tab:city-speed-comp}.

%% file: conclusion.tex
\section{Conclusion}
In this paper, we present a novel knowledge distill framework tailored for semantic segmentation. We improve the performance of the student model by translating the high-level feature to a compact format that is easier to be learned.
Extensive experiments have been done to testify the effectiveness of our proposed method.
Even built upon a highly competitive baseline, our method (1) improves the performance of student model by a large margin without introducing extra parameters or computations (2) achieves better results with much less computation overheads.

%% file: supplymaterial.tex
\section*{Appendix}

\section{Implementation details}
\textbf{Knowledge Translator:} In all our experiments, the auto-encoder is composed of three 2-D convolution layers and three 2-D transposed convolution layers.
The convolution strides of the first convolution layer in the encoder and the last convolution layer in the decoder are set to 2. All six layers use a $3\times3$ kernel with padding of 1, BN layer and ReLU activation function. The channels of the six convolution layers are all equal to the channels of the last feature maps in the teacher model.

\textbf{Knowledge Adapter:} The adapter is utilized on the top of the last convolution features of the student model. It contains three convolution layers using a $3\times3$ kernel with stride of 1, padding of 1, BN layer and ReLU activation function. The spatial resolution of the feature maps of the three convolution layers remain unchanged and the channels are adjusted to the number of the features in the last layer of the teacher model.

\textbf{Training Process:}
We follow the training strategy of DeepLabV3+ \cite{Chen2018deeplabv3plus} and MobileNetV2 \cite{Sandlermobilenetv22018} to train the teacher and the student models, respectively.
For the Pascal VOC dataset, we firstly train the teacher and the student networks for 500,000 iterations on the COCO dataset. 30,000 iterations are followed by training on the \textit{trainaug} dataset. We validate the performance on the \textit{val} set.
For the Pascal Context dataset, we train our teacher and student model on the \textit{train} set for 30,000 iterations and the performance is tested on the \textit{val} set.
For the Cityscapes dataset, we train our model for 90,000 iterations on the \textit{train-fine} dataset, which is fine tuned on the \textit{trainval} and the \textit{train-coarse} sets for another 90,000 iterations and the performance is evaluated on the \textit{test} set. Our reported model is not pre-trained on the COCO dataset.
We set the learning rate to 0.007 and the total batch size of 64 in all our experiments. We train our model by using 4 GPUs with crop size of $513\times513$ on the Pascal VOC and Pascal Context and $769\times769$ on the Cityscapes.

\section{Results Visualization}
\begin{figure*}
	\begin{center}
		\includegraphics[width=15cm]{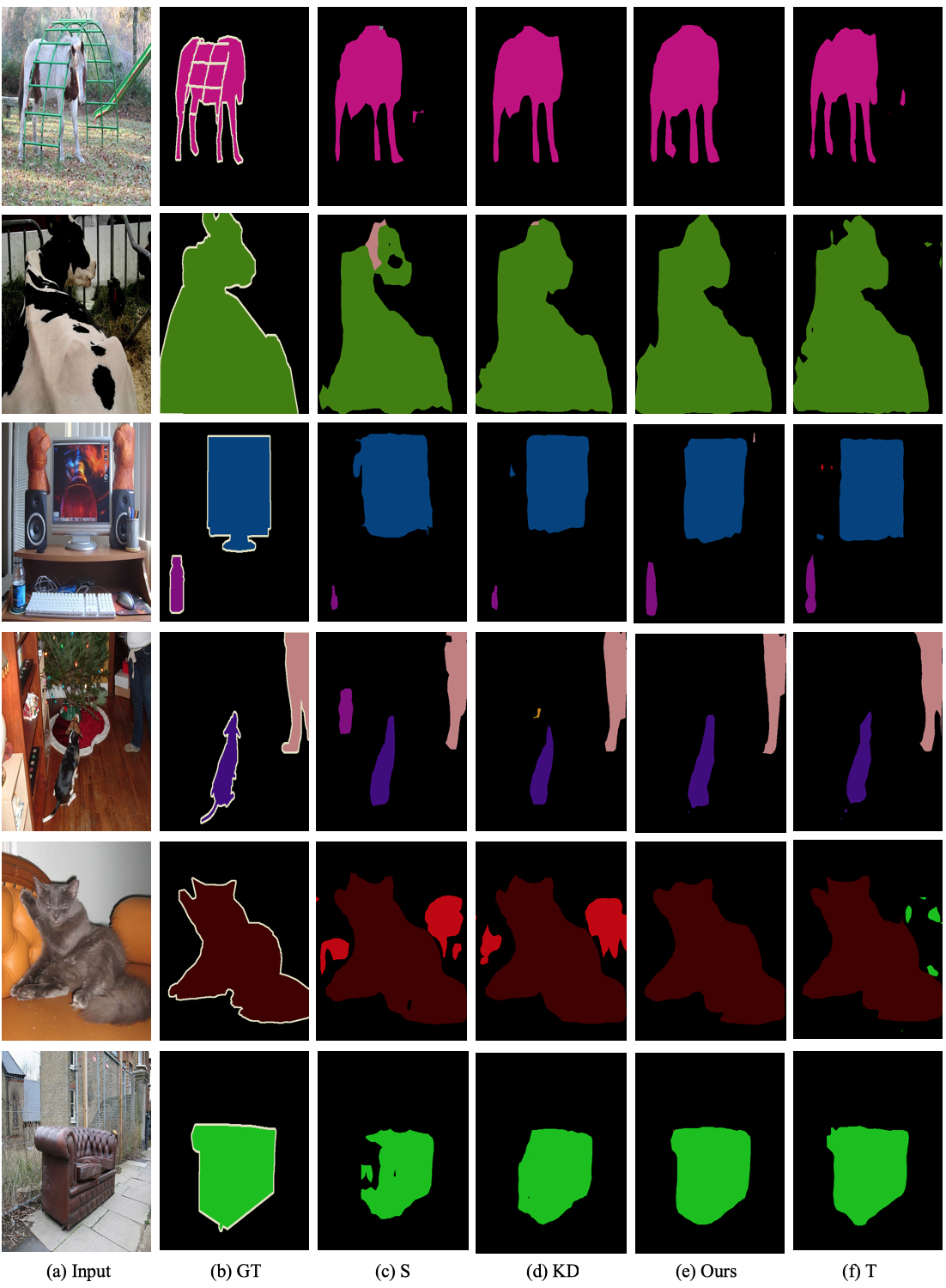}
	\end{center}
	\caption{Comparison of segmentation results on Pascal VOC and Pascal Context datasets. (a) Input image. (b) Ground truth. (c) The results of the student network (MobileNetV2). (d) The results of the knowledge distillation \cite{Hinton2015distill}. (e) The results of our proposed method. (f) The results of the teacher network ResNet50.}
	\label{fig:supp_results}
\end{figure*}

\begin{figure*}
	\begin{center}
		\includegraphics[width=17cm]{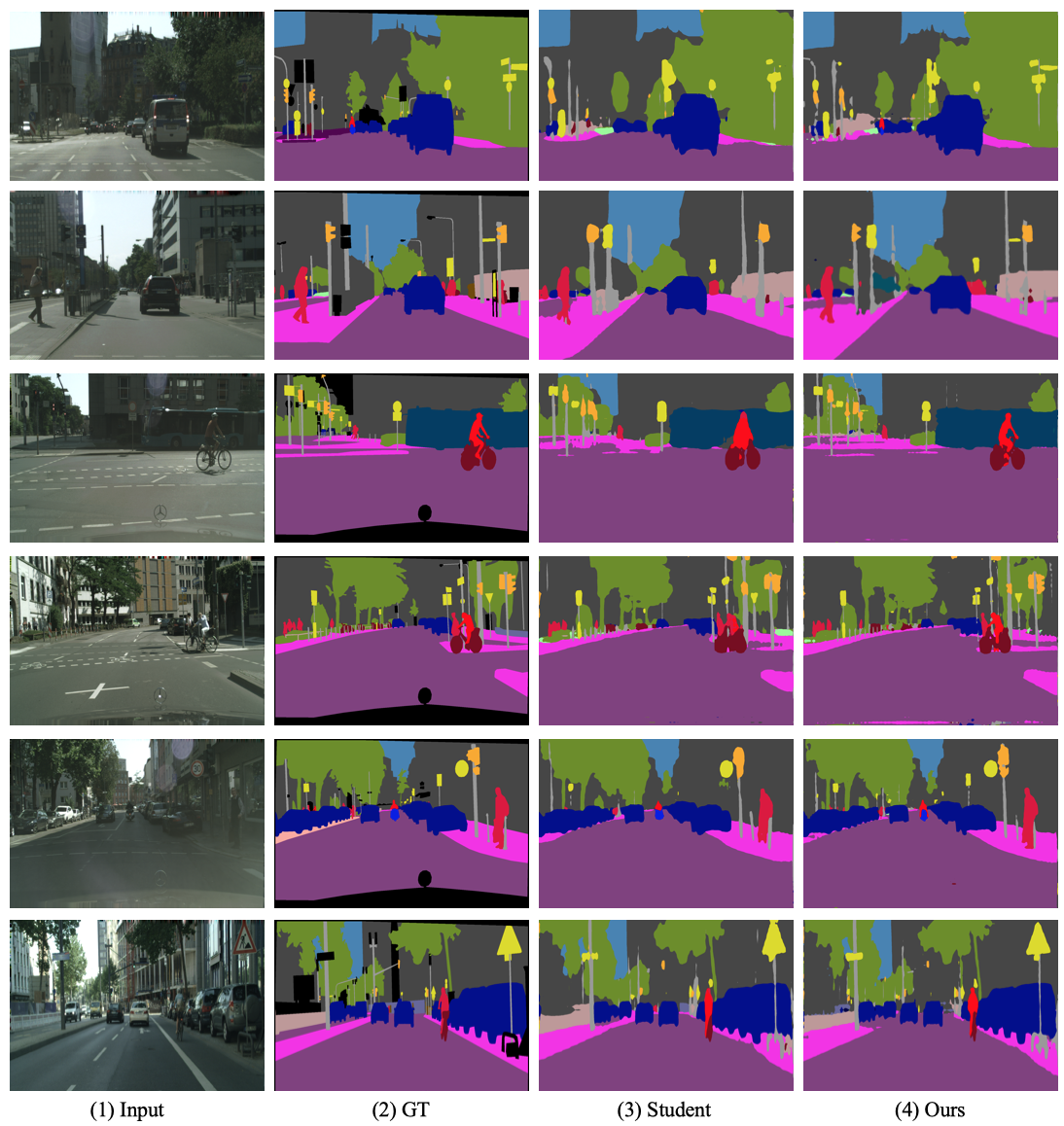}
	\end{center}
	\vspace{-0.1cm}
	\caption{Example results on the Cityscapes dataset. From left to right are: (1) Input images, (2) Ground truth, (3) The results of the student net (4) The results of our proposed method.}
	\label{fig:supp_cityscapes}
\end{figure*}

We first visualize our results on the Pascal VOC dataset. Our method is built upon a strong baseline with the student net of MobileNetV2 \cite{Sandlermobilenetv22018}, which does not contain the ASPP \cite{Chen2018deeplabv3plus} or the decoder. As shown in Figure \ref{fig:supp_results}, our method generates more accurate segmentation results than the student network and also outperforms the results of \cite{Hinton2015distill}.
To further demonstrate the effectiveness of our proposed method, more results on the Cityscapes dataset are shown in Figure \ref{fig:supp_cityscapes}.

%% file: final.bbl
\begin{thebibliography}{10}\itemsep=-1pt

\bibitem{Ayinde2018deep}
B.~O. Ainde and J.~M. Zurada.
\newblock {Deep Learning of Constrained Autoencoders for Enhanced Understanding
  of Data}.
\newblock 2018.

\bibitem{Badrinarayanan2017segnet}
V.~Badrinarayanan, A.~Kendall, and R.~Cipolla.
\newblock {SegNet: A Deep Convolutional Encoder-Decoder Architecture For Image
  Segmentation}.
\newblock {\em {IEEE} Trans. Pattern Anal. Mach. Intell.}, 2017.

\bibitem{Chen2017deeplab}
L.-C. Chen, G.~Papandreou, I.~Kokkinos, K.~Murphy, and A.~Yuille.
\newblock {DeepLab: Semantic Image Segmentation With Deep Convolutional Nets,
  Atrous Convolution, And Fully Connected Crfs}.
\newblock {\em {IEEE} Trans. Pattern Anal. Mach. Intell.}, 2017.

\bibitem{Chen2017deeplabv3}
L.-C. Chen, G.~Papandreou, F.~Schroff, and H.~Adam.
\newblock {Rethinking Atrous Convolution for Semantic Image Segmentation}.
\newblock {\em arXiv preprint arXiv:1706.05587}, 2017.

\bibitem{Chen2018deeplabv3plus}
L.-C. Chen, Y.~Zhu, G.~Papandreou, F.~Schroff, and H.~Adam.
\newblock {Encoder-Decoder with Atrous Separable Convolution for Semantic Image
  Segmentation}.
\newblock In {\em Proc. Eur. Conf. Comp. Vis.}, 2018.

\bibitem{Chollet2017xception}
F.~Chollet.
\newblock {Deep Learning With Depthwise Separable Convolutions}.
\newblock In {\em Proc. IEEE Conf. Comp. Vis. Patt. Recogn.}, 2017.

\bibitem{Cordts2016Cityscapes}
M.~Cordts, M.~Omran, S.~Ramos, T.~Rehfeld, M.~Enzweiler, R.~Benenson,
  U.~Franke, S.~Roth, and B.~Schiele.
\newblock {The Cityscapes Dataset For Semantic Urban Scene Understanding}.
\newblock In {\em Proc. IEEE Conf. Comp. Vis. Patt. Recogn.}, 2016.

\bibitem{Everingham2014pascal}
M.~Everingham, S.~Eslami, L.~Gool, C.~Williams, J.~Winn, and A.~Zisserman.
\newblock {The Pascal Visual Object Classes Challenge – A Retrospective}.
\newblock {\em Int. J. Comput. Vision}, 2014.

\bibitem{Hariharan2011semantic}
B.~Hariharan, P.~Arbelaez, L.~Bourdev, S.~Maji, and J.~Malik.
\newblock {Semantic Contours From Inverse Detectors}.
\newblock In {\em Proc. IEEE Conf. Comp. Vis. Patt. Recogn.}, 2011.

\bibitem{He2015residual}
K.~He, X.~Zhang, S.~Ren, and J.~Sun.
\newblock {Deep Residual Learning For Image Recognition}.
\newblock In {\em Proc. IEEE Conf. Comp. Vis. Patt. Recogn.}, 2016.

\bibitem{Hinton2015distill}
G.~Hinton, O.~Vinyals, and J.~Dean.
\newblock {Distilling the Knowledge in a Neural Network}.
\newblock {\em arXiv preprint arXiv:1503.02531}.

\bibitem{Huang2017like}
Z.~Huang and N.~Wang.
\newblock {Like What You Like: Knowledge Distill via Neuron Selectivity
  Transfer}.
\newblock {\em arXiv preprint arXiv:1707.01219}, 2017.

\bibitem{Jangho2018paraphrasing}
J.~Kim, S.~Park, and N.~Kwak.
\newblock {Paraphrasing Complex Network: Network Compression via Factor
  Transfer}.
\newblock In {\em Proc. Advances in Neural Inf. Process. Syst.}, 2018.

\bibitem{Lin2017refinenet}
G.~Lin, A.~Milan, C.~Shen, and I.~Reid.
\newblock {RefineNet}: Multi-path refinement networks for high-resolution
  semantic segmentation.
\newblock In {\em Proc. IEEE Conf. Comp. Vis. Patt. Recogn.}, 2017.

\bibitem{Lin2015efficient}
G.~Lin, C.~Shen, A.~van~den Hengel, and I.~Reid.
\newblock {Efficient piecewise training of deep structured models for semantic
  segmentation}.
\newblock In {\em Proc. IEEE Conf. Comp. Vis. Patt. Recogn.}, 2016.

\bibitem{Liu2015semantic}
Z.~Liu, X.~Li, P.~Luo, C.~Loy, and X.~Tang.
\newblock {Semantic Image Segmentation Via Deep Parsing Network}.
\newblock In {\em Proc. IEEE Int. Conf. Comp. Vis.}, 2015.

\bibitem{Long2015fully}
J.~Long, E.~Shelhamer, and T.~Darrell.
\newblock {Fully Convolutional Networks For Semantic Segmentation}.
\newblock In {\em Proc. IEEE Conf. Comp. Vis. Patt. Recogn.}, 2015.

\bibitem{Mottaghi2014}
R.~Mottaghi, X.~Chen, X.~Liu, N.~Cho, S.~Lee, S.~Fidler, R.~Urtasun, and
  A.~Yuille.
\newblock {The Role of Context for Object Detection and Semantic Segmentation
  in the Wild}.
\newblock In {\em Proc. IEEE Conf. Comp. Vis. Patt. Recogn.}, 2014.

\bibitem{Paszke2016enet}
A.~Paszke, A.~Chaurasia, S.~Kim, and E.~Culurciello.
\newblock {ENet: A Deep Neural Network Architecture for Real-Time Semantic
  Segmentation }.
\newblock {\em arXiv preprint arXiv:1606.02147}, 2016.

\bibitem{Romero2015fitnets}
A.~Romero, N.~Ballas, S.~Kahou, A.~Chassang, C.~Gatta, and Y.~Bengio.
\newblock {FitNets: Hints for Thin Deep Nets}.
\newblock {\em arXiv preprint arXiv:1412.6550}.

\bibitem{Sandlermobilenetv22018}
M.~Sandler, A.~Howard, M.~Zhu, A.~Zhmoginov, and L.-C. Chen.
\newblock {MobileNetV2: Inverted Residuals and Linear Bottlenecks}.
\newblock In {\em CVPR}, 2018.

\bibitem{Shuai2017dag}
B.~Shuai, Z.~Zuo, B.~Wang, and G.~Wang.
\newblock {Scene Segmentation With Dagrecurrent Neural Networks}.
\newblock {\em {IEEE} Trans. Pattern Anal. Mach. Intell.}, 2017.

\bibitem{Simonyan2014vggnet}
K.~Simonyan and A.~Zisserman.
\newblock {Very Deep Convolutional Networks for Large-Scale Image Recognition
  }.
\newblock {\em arXiv preprint arXiv:1409.1556}.

\bibitem{Treml2016sq}
M.~Treml, J.~Arjona-Medina, T.~Unterthiner, R.~Durgesh, F.~Friedmann,
  P.~Schuberth, A.~Mayr, M.~Heusel, M.~Hofmarcher, and M.~W. et~al.
\newblock {Speeding Up Semantic Segmentation For Autonomous Driving }.
\newblock 2016.

\bibitem{Wang2018nonlocal}
X.~Wang, R.~Girshick, A.~Gupta, and K.~He.
\newblock {Non-local Neural Networks}.
\newblock In {\em CVPR}, 2018.

\bibitem{Wu2017resnet38}
Z.~Wu, C.~Shen, and A.~van~den Hengel.
\newblock {Wider or Deeper: Revisiting the ResNet Model for Visual
  Recognition}.
\newblock {\em arXiv preprint arXiv:1611.10080}, 2016.

\bibitem{Yim2017agift}
J.~Yim, D.~Joo, J.~Bae, and J.~Kim.
\newblock {A Gift from Knowledge Distillation: Fast Optimization, Network
  Minimization and Transfer Learning}.
\newblock In {\em Proc. IEEE Conf. Comp. Vis. Patt. Recogn.}, 2017.

\bibitem{Yu2015multiscale}
F.~Yu and V.~Koltun.
\newblock {Multi-Scale Context Aggregation by Dilated Convolutions}.
\newblock {\em arXiv preprint arXiv:1511.07122}.

\bibitem{Zagoruyko2017atnet}
S.~Zagoruyko and N.~Komodakis.
\newblock {Paying More Attention to Attention: Improving the Performance of
  Convolutional Neural Networks via Attention Transfer}.
\newblock In {\em Proc. Int. Conf. Learn. Representations}, 2017.

\bibitem{Zhang2018exfuse}
Z.~Zhang, X.~Zhang, C.~Peng, D.~Cheng, and J.~Sun.
\newblock {ExFuse: Enhancing Feature Fusion for Semantic Segmentation}.
\newblock In {\em Proc. Eur. Conf. Comp. Vis.}, 2018.

\bibitem{Zhao2017icnet}
H.~Zhao, X.~Qi, X.~Shen, J.~Shi, and J.~Jia.
\newblock {ICNet for Real-Time Semantic Segmentation on High-Resolution
  Images}.
\newblock {\em arXiv preprint arXiv:1704.08545}, 2017.

\bibitem{Zhao2017psp}
H.~Zhao, J.~Shi, X.~Qi, X.~Wang, and J.~Jia.
\newblock {Pyramid Scene Parsing Network}.
\newblock In {\em Proc. IEEE Conf. Comp. Vis. Patt. Recogn.}, 2017.

\bibitem{Zheng2015crfasrnn}
S.~Zheng, S.~Jayasumana, B.~Romera-Paredes, V.~Vineet, Z.~Su, D.~Du, C.~Huang,
  and P.~Torr.
\newblock {Conditional Random Fields as Recurrent Neural Networks}.
\newblock In {\em Proc. IEEE Int. Conf. Comp. Vis.}, 2015.

\end{thebibliography}
